\documentclass[10pt,twocolumn,letterpaper]{article}

\usepackage[algorithms]{wacv} 

\usepackage{booktabs}
\usepackage{multirow}
\usepackage{xspace}
\usepackage{amssymb}
\usepackage{amsmath}
\usepackage{tikz}
\usetikzlibrary{positioning,calc,arrows.meta}
\usepackage{colortbl}
\usepackage[accsupp]{axessibility} 

\definecolor{wacvblue}{rgb}{0.21,0.49,0.74}
\usepackage[pagebackref,breaklinks,colorlinks,allcolors=wacvblue]{hyperref}

\def\wacvPaperID{7}
\def\confName{WACV}
\def\confYear{2027}

\newcommand{\ours}{SaV\xspace}            
\newcommand{\RFS}{RFS\xspace}

\newcommand{\valpool}{val-1.9K\xspace}
\newcommand{\valr}{val-456\xspace}
\newcommand{\dev}{dev-630\xspace}
\definecolor{oursrow}{RGB}{253,234,234}
\newcommand{\oursrow}{\rowcolor{oursrow}}
\newcommand{\hl}[1]{\textcolor{red!80!black}{\textbf{#1}}}

\title{Greedy Decoding Hides Good Plans:\\Sample-and-Verify Planning for Vision-Language Driving Models}

\author{Zhipeng Bao \quad Wenjie Zhao \quad Yiping Li \quad Haohua Que \quad Qianwen Li$^{*}$\\
University of Georgia, United States\\
{\tt\small zb28097@uga.edu, Cami.Li@uga.edu}\\
{\small $^{*}$Corresponding author}
}

\begin{document}
\twocolumn[{%
\renewcommand\twocolumn[1][]{#1}%
\maketitle
\begin{center}
\vspace{-7mm}
\includegraphics[width=\textwidth]{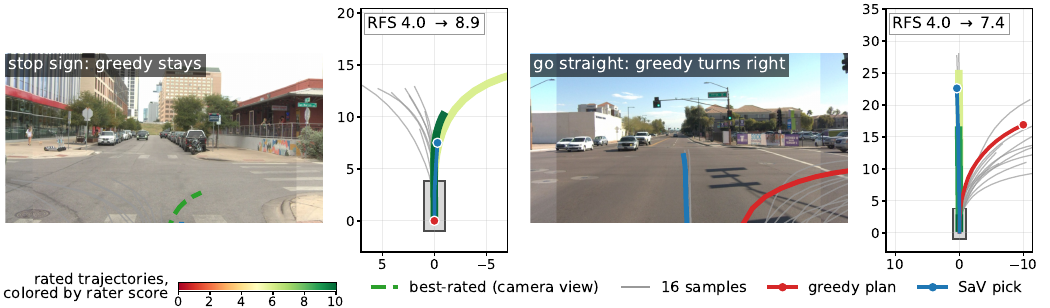}
\vspace{-5mm}
\captionof{figure}{\textbf{Greedy decoding hides good plans.}
Camera views and bird's-eye-view trajectories illustrate two cases where better plans are already present among the samples.
\emph{Left:} greedy decoding stays while \ours selects a starting trajectory preferred by the raters (\RFS 4.0$\to$8.9).
\emph{Right:} greedy decoding turns right although the preferred behavior is to proceed straight; \ours selects a straight trajectory (\RFS 4.0$\to$7.4).}
\label{fig:teaser}
\end{center}
\vspace{3mm}
}]
\begin{abstract}
Vision-language models (VLMs) are increasingly adopted for end-to-end driving planning, yet their performance is typically evaluated based on a single greedily decoded trajectory. However, greedy decoding may overlook better trajectories within the model's output distribution. Experiments on WOD-E2E and NAVSIM reveal a consistent performance gap between greedily decoded trajectories and the best sampled candidates, showing that planning errors can arise from trajectory selection, rather than trajectory generation alone. On WOD-E2E, when the ego vehicle should start moving, the model often assigns probability to different possible start times, causing greedy decoding to repeatedly favor waiting and produce a stationary plan even though moving trajectories are available. Motivated by these findings, we propose Sample-and-Verify (SaV), which keeps the VLM planner frozen and samples multiple candidate trajectories. A lightweight verifier is trained from ground-truth futures to estimate the relative quality of these candidates. It uses trajectory kinematics, candidate-set agreement and likelihood, and ego-motion features to select the final trajectory. On WOD-E2E, SaV reduces ADE and FDE by approximately 15\% across three VLM planners on the validation set. Changes in RFS are smaller: two planners show slight decreases on val-456, while Alpamayo-1.5 improves from 7.74 to 7.81 RFS on the official test set as ADE@5s decreases by 11\%. On NAVSIM, SaV reduces ADE@4s by 10--24\% across three VLM planners, while PDMS changes are less consistent. These results show that trajectory selection can substantially improve displacement accuracy without additional planner training. The code is publicly available at \url{https://github.com/hzxbzp/sample-and-verify}.
\end{abstract}

\section{Introduction}
\label{sec:intro}

Vision-language models (VLMs) are increasingly used as end-to-end driving planners, mapping visual observations and ego states to future trajectories, often together with intermediate reasoning~\cite{hwang2024emma,wang2025alpamayo,rowe2025poutine,zhou2026autovla}. In several recent approaches, trajectory planning is formulated autoregressively by representing future motion as sequences of textual coordinates or discrete action tokens~\cite{hwang2024emma,rowe2025poutine,zhou2026autovla}. Research has largely focused on improving what these models learn through stronger visual reasoning and grounding, better action representations, additional driving supervision, or reinforcement-learning post-training~\cite{sima2024drivelm,bao2026anchorreasoning,zheng2026drivema,rowe2025poutine,zhou2026autovla}.

How these planners are decoded at inference time, however, has received less attention. An autoregressive planner learns a distribution over possible future trajectories, but inference requires decoding a single trajectory from that distribution. Greedy decoding makes this selection locally, choosing the most likely token at each generation step. Choosing the most likely token at each step does not necessarily produce the best trajectory overall, especially when the model assigns probability to several plausible behaviors or different timings of the same behavior. Consequently, the quality of a greedily decoded trajectory may not faithfully reflect the planning capability represented by the model distribution.

This distinction raises a fundamental question:
\emph{When a VLM produces a poor plan, does the model fail to contain an appropriate trajectory, or does its decoding procedure fail to recover one that is already represented in its output distribution?} These two failure modes call for fundamentally different solutions. The former motivates additional data, supervision, or model training. The latter is an inference-time search and selection problem. Existing evaluations, which typically judge only the final decoded trajectory, do not distinguish between them.

We study this question on the Waymo Open Dataset for End-to-End Driving (WOD-E2E)~\cite{xu2026wode2e} and NAVSIM~\cite{dauner2024navsim}. Across both benchmarks, we find a gap between the trajectories selected by greedy decoding and the better alternatives available in the same model distribution. A prominent example occurs when the ego vehicle is stationary and should proceed: greedy trajectories from three backbones frequently remain stopped, while alternative moving trajectories remain present in the model distribution. We trace this failure to uncertainty in start time, which causes greedy decoding to repeatedly favor waiting. More broadly, the results on both benchmarks show that greedy decoding can fail to recover better trajectories already generated by the same planner.

Motivated by this observation, we introduce \emph{Sample-and-Verify} (\ours), a test-time planning framework that separates trajectory generation from trajectory selection. The frozen planner first generates its greedy trajectory together with $K$ sampled candidates. A lightweight verifier then scores each candidate using its kinematics, agreement with the candidate set, model likelihood, and the ego-motion history, and selects the highest-quality plan. Experiments show that \ours consistently improves trajectory accuracy. On WOD-E2E, \ours reduces 5-s average displacement error (ADE) and final displacement error (FDE) by approximately 15--16\% across three VLM planners, reduces failed starts, and transfers across VLMs without retraining. On the official WOD-E2E leaderboard, it improves Alpamayo-1.5's \RFS from 7.74 to 7.81 while reducing ADE@5s by 11\%. On NAVSIM, \ours reduces ADE@4s by 9--25\% across three publicly released planners. Together, these results suggest that decoding can affect the planning performance of autoregressive VLM-based driving models.

\noindent Our contributions are:
\begin{itemize}
    \item We identify trajectory selection as a distinct failure source in autoregressive VLM driving planners, showing that greedy decoding can miss better trajectories already present in the model distribution. This distinction separates decoding failures from trajectory-generation failures and motivates evaluating them independently.

    \item We introduce SaV, a test-time framework that samples candidate trajectories from a frozen VLM planner and ranks them with a lightweight supervised verifier. The planner requires no additional training; only the verifier is trained from logged ground-truth futures.
\end{itemize}

\section{Related Work}
\label{sec:related}

\noindent\textbf{Vision-language models for driving.}
Early vision-language driving work focused on explanation, question answering, and structured reasoning~\cite{kim2018bddx,malla2023drama,qian2024nuscenesqa,marcu2024lingoqa,sachdeva2024rank2tell,sima2024drivelm}. Recent VLM/VLA approaches increasingly generate driving actions or future trajectories directly~\cite{hwang2024emma,tian2024drivevlm,chi2026impromptu,zhou2026autovla,wang2025alpamayo}, with further improvements from reinforcement-learning post-training and alternative action representations~\cite{rowe2025poutine,zheng2026drivema}.
Unlike these works, we keep the trained VLM fixed and study how its trajectory distribution is decoded at inference time.

\noindent\textbf{Reasoning and grounding for driving.}
Recent methods introduce chain-of-thought and visual grounding to connect scene observations with driving decisions~\cite{nie2024reason2drive,ishaq2025drivelmm,huang2026nureasoning,bao2026anchorreasoning}. Other work improves reasoning reliability through deferred trajectory exposure or verifiable meta-actions~\cite{huang2026deft,zheng2026drivema}. \ours is complementary to these approaches: it does not change what the planner learns, but selects among trajectories the planner already generates.

\noindent\textbf{Decoding and trajectory selection.}
Sequence models can assign poor outputs high probability, motivating sampling-based methods such as minimum-Bayes-risk decoding~\cite{stahlberg2019cat,eikema2020map,holtzman2020degeneration} and test-time verification~\cite{wang2023selfconsistency,cobbe2021verifiers,lightman2024verify,ma2026cyclevla}. Autonomous driving methods also generate and score multiple trajectory candidates~\cite{chen2026vadv2,li2024hydramdp,li2025gtrs,yao2025drivesuprim,guo2025ipad}.
Our setting differs in that the candidates are sampled directly from the same autoregressive VLM planner, and the verifier is trained only from logged futures. Behavior cloning is also known to exhibit inertia when a stopped vehicle tends to remain stopped~\cite{codevilla2019limitations,wen2020copycat}. We show that greedy decoding can contribute to this failure even when sampled trajectories contain moving plans.
\section{Why Greedy Plans Fail}
\label{sec:analysis}

\noindent\textbf{Setting.}
We study three VLM planners with different model scales and pre-training: Cosmos-Reason2-2B~\cite{nvidia2025cosmosreason2}, Qwen3-VL-8B~\cite{bai2025qwen3vl}, and the driving-specialized Alpamayo-1.5-10B~\cite{wang2025alpamayo}. Each model is fine-tuned from the AnchorReasoning grounding stage~\cite{bao2026anchorreasoning} to predict a 5\,s trajectory from the current three-camera panorama and a 4\,s ego-motion history. Greedy decoding gives \RFS scores of 7.56, 7.54, and 7.81 on the WOD-E2E validation frames (val-456). We define the ego as \emph{stationary} when its speed is below 0.5\,m/s, a plan \emph{starts} when its 5\,s path length exceeds 3\,m and \emph{stays} when it is below 1\,m.

\begin{table}[t]
\centering
\small
\setlength{\tabcolsep}{3.5pt}
\resizebox{\columnwidth}{!}{%
\begin{tabular}{llcccc}
\toprule
Model & Method & FTS (\%)$\downarrow$ & FSR (\%)$\downarrow$ &
Stop sign$\downarrow$ & \RFS$\uparrow$\\
\midrule
Cosmos-Reason2
& Greedy & 84 & 5  & 12/12 & 6.73\\
(2B)
& SaV    & 17 & 32 & 3/12  & 6.94\\

Qwen3-VL
& Greedy & 41 & 14 & 4/12  & 6.64\\
(8B)
& SaV    & 21 & 36 & 1/12  & 6.65\\

Alpamayo-1.5
& Greedy & 43 & 9  & 6/12  & 6.57\\
(10B)
& SaV    & 30 & 27 & 5/12  & 6.81\\
\bottomrule
\end{tabular}}
\caption{\textbf{Failed and false starts on the rated validation frames.}
Failed-start rate (FTS) is evaluated on the 82 frames where the ego is stationary and the best-rated trajectory starts, while false-start rate (FSR) is evaluated on the 22 frames where the best-rated trajectory stays. The stop-sign column reports failed starts among the 12 stop-sign cases. \RFS is reported on the 82 starting frames.}
\label{tab:fts}
\end{table}

\noindent\textbf{Greedy decoding may collapse uncertainty in start time.}
A common failure occurs when the ego vehicle is stationary but should begin moving.
Among the 456 rated validation frames, 82 fall into this case.
The three models fail to start in 41--84\% of these frames and remain stopped in 4--12 of the 12 stop-sign cases (\cref{tab:fts}).

\begin{figure*}[t]
\centering
\includegraphics[width=0.92\textwidth]{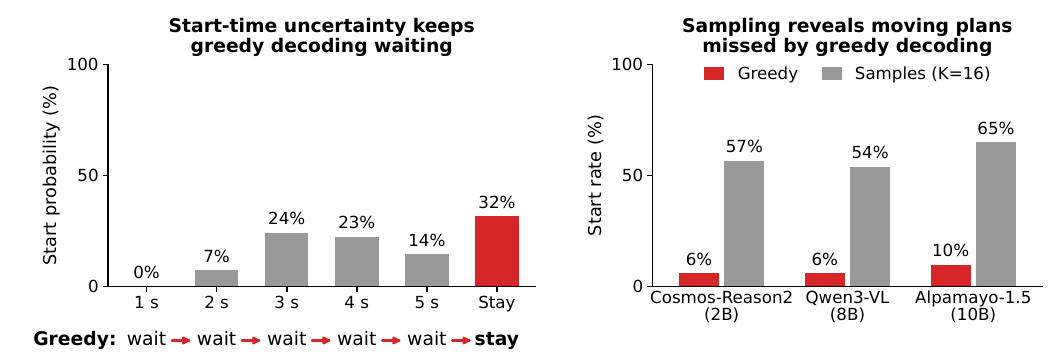}
\caption{\textbf{Start-time uncertainty is associated with failed starts under greedy decoding.}
Left: empirical distribution of ground-truth start times in stationary stop-sign training frames, showing substantial variation in when motion begins. Right: start rates of greedy and sampled trajectories on validation frames where the ego is stationary and the ground-truth trajectory starts. The aggregate start-time distribution motivates the hypothesis that uncertainty in start timing contributes to greedy waiting.}
\label{fig:collapse}
\end{figure*}

This behavior cannot be explained simply by insufficient starting examples in the training data.
Among stationary training frames, 43\% of the ground-truth trajectories start within 5\,s, increasing to 68\% at stop signs.
Instead, the start probability is distributed across different future times (\cref{fig:collapse}, left).
At stop signs, some trajectories start early while others wait longer, so no single start time dominates the distribution.
Greedy decoding selects the most likely continuation at each step; repeatedly favoring ``wait'' can therefore produce a staying trajectory even though many ground-truth trajectories eventually start.
Consistent with this analysis, the trained models start in only 6--10\% of the corresponding validation cases under greedy decoding (\cref{fig:collapse}, right).

A cross-dataset check on NAVSIM further supports this interpretation that the broader gap between greedy decoding and better candidate trajectories remains (\cref{sec:main_results}).

\noindent\textbf{Consensus alone is insufficient.}
Sampling provides diverse motion hypotheses, but simply selecting the consensus trajectory does not reliably recover the appropriate plan. In the validation frames where the ground truth remains stationary, 40--42\% of sampled trajectories still start.
Accordingly, minimum-Bayes-risk (MBR) decoding reduces failed starts but introduces false starts and lowers \RFS across all three models (\cref{tab:consensus}). The sample mean and majority-start rule show the same limitation.
Thus, sampling exposes alternative trajectories, but selecting an appropriate candidate remains a separate problem.

\begin{table}[t]
\centering
\small
\setlength{\tabcolsep}{4pt}
\begin{tabular}{lccc}
\toprule
Strategy & Cosmos & Qwen3 & Alpamayo \\
\midrule
Greedy         & 7.56 & 7.54 & 7.81 \\
Sample mean    & 7.28 & 7.25 & 7.55 \\
MBR selection  & 7.47 & 7.39 & 7.73 \\
Majority-start & 7.48 & 7.43 & 7.79 \\
\midrule
GT             & 8.88 & 8.86 & 9.18 \\
\bottomrule
\end{tabular}
\caption{\textbf{RFS comparison of different sampling and selection strategies on the rated validation frames.}
Each sampling-based method uses 16 trajectories per frame, while greedy decoding serves as the baseline.
GT reports the highest \RFS trajectory among the rated candidates offered by WOD-E2E.}
\label{tab:consensus}
\end{table}

\section{Sample-and-Verify Planning}
\label{sec:method}

\begin{figure}[t]
    \centering
    \includegraphics[width=\columnwidth]{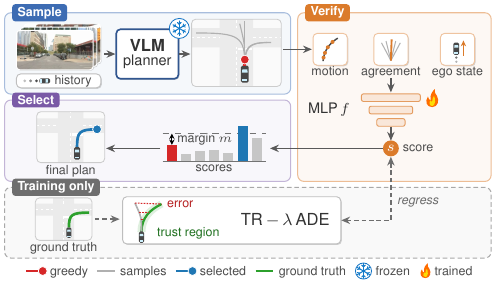}
    \caption{\textbf{Overview of Sample-and-Verify planning.}
During training, candidate quality combines a trust-region (TR) score that measures consistency with the ground-truth trajectory and average displacement error (ADE) that measures geometric deviation from it.}
    \label{fig:method}
\end{figure}

\ours addresses failures caused by selecting a single trajectory from the output distribution of a VLM planner. As illustrated in \cref{fig:method}, \ours keeps the fine-tuned planner frozen and separates planning into two stages. The planner first generates multiple candidate trajectories, and a lightweight verifier then evaluates these candidates and selects one for execution.

\noindent\textbf{Candidate generation.}
Given an input $x$, the frozen planner produces a greedy trajectory $\tau_0$ and $K$ sampled trajectories $\tau_1,\ldots,\tau_K$.
We denote the complete candidate set as
\begin{equation}
\mathcal{T}=\{\tau_0,\tau_1,\ldots,\tau_K\}.
\label{eq:candidate_set}
\end{equation}
For each candidate $\tau_k$, we also retain its log-likelihood $\ell_k$ under the planner.

\noindent\textbf{Verifier representation.}
For each candidate $\tau_k \in \mathcal{T}$, the verifier constructs a feature representation
\begin{equation}
\phi_k = \phi(\tau_k \mid x,\mathcal{T}),
\label{eq:features}
\end{equation}
which combines three types of information.
\emph{Trajectory features} describe the motion and kinematic properties of the candidate, including displacement, speed, acceleration, jerk, heading, and motion initiation.
\emph{Candidate-set features} characterize the candidate relative to the other trajectories in $\mathcal{T}$, including its agreement with the sampled trajectories, deviation from the greedy trajectory, relative path length, and planner likelihood.
\emph{Ego-state features} describe the recent ego motion and the navigation command.
Together, these features allow the verifier to assess both the properties of an individual trajectory and its relation to the current ego state and alternative plans.

\noindent\textbf{Verifier target.}
We construct a trajectory-quality target from the future trajectory ground-truth $\tau^\ast$. For each candidate $\tau_k$, we define
\begin{equation}
y_k =
\mathrm{TR}(\tau_k;\tau^\ast,v_0)
-
\lambda\,\mathrm{ADE}_{5\mathrm{s}}(\tau_k,\tau^\ast),
\label{eq:target}
\end{equation}
where $\mathrm{TR}(\tau_k;\tau^\ast,v_0)$ denotes the trust-region score between the candidate $\tau_k$ and the $\tau^\ast$, with the tolerance determined by the initial ego speed $v_0$. The trust-region term follows the structure of the official \RFS metric and tolerates speed-dependent deviations from the reference trajectory.
The ADE term further penalizes geometric deviation from the ground-truth, while $\lambda$ controls its contribution to the training target.

Because the verifier only needs to distinguish the quality of candidates within the same scene, we train it to predict relative rather than absolute trajectory quality.

\noindent\textbf{Verifier training.}
The verifier $f$ is a lightweight multilayer perceptron (MLP) with two hidden layers of 128 units each.
Given the feature representation $\phi_k$, it predicts a scalar quality score
\begin{equation}
s_k = f(\phi_k).
\label{eq:verifier_score}
\end{equation}
Because trajectory selection depends on the relative quality of candidates within the same scene rather than their absolute scores across scenes, we train the verifier to predict centered candidate quality.
The training objective is
\begin{equation}
\mathcal{L}
=
\frac{1}{K+1}
\sum_{k=0}^{K}
\left[
(s_k-\bar{s})-(y_k-\bar{y})
\right]^2,
\label{eq:loss}
\end{equation}
where
\begin{equation}
\bar{s}
=
\frac{1}{K+1}\sum_{k=0}^{K}s_k,
\qquad
\bar{y}
=
\frac{1}{K+1}\sum_{k=0}^{K}y_k.
\end{equation}
Centering the predicted and target scores within each scene removes scene-level score offsets and focuses the verifier on differences in quality among the candidate trajectories.

\noindent\textbf{Trajectory selection.}
At test time, the verifier scores every trajectory in $\mathcal{T}$ and identifies the highest-scoring candidate:
\begin{equation}
k^\ast =
\arg\max_{k\in\{0,\ldots,K\}} s_k.
\label{eq:best_candidate}
\end{equation}
To avoid replacing the greedy trajectory when the verifier shows only a weak preference for an alternative, \ours uses a positive selection margin $m>0$:
\begin{equation}
\tau_{\mathrm{out}}
=
\begin{cases}
\tau_{k^\ast},
& s_{k^\ast}-s_0 \ge m,\\
\tau_0,
& \text{otherwise}.
\end{cases}
\label{eq:selection}
\end{equation}
Thus, \ours does not modify the VLM planner or its training objective.
Instead, it operates at test time by selecting among trajectories already generated by the frozen planner.
\section{Experiments}
\label{sec:experiments}

\noindent\textbf{Data and metrics.}
We evaluate on WOD-E2E~\cite{xu2026wode2e} and NAVSIM~\cite{dauner2024navsim}.
For WOD-E2E, \valr contains 456 frames with human-rated trajectories for \RFS evaluation, while \valpool contains 1,916 additional validation frames evaluated against the ground-truth future using ADE and FDE at 5\,s.
We use \dev (630 held-out training frames) for verifier validation and margin selection.
For NAVSIM, we evaluate Curious-VLA-SFT, Curious-VLA-RL~\cite{chen2026curious}, and ReCogDrive-VLM-8B~\cite{li2026recogdrive} on all 12,146 navtest scenes using the official PDM score (PDMS) and ADE at 4\,s.
Their verifiers are trained on 7,989 navtrain scenes.
For both benchmarks, we also compare greedy decoding with alternative sampling and selection strategies.

\noindent\textbf{Implementation details.}
On WOD-E2E, we evaluate Cosmos-Reason2 (2B), Qwen3-VL (8B), and Alpamayo-1.5 (10B), fine-tuned from their AnchorReasoning grounding-stage checkpoints~\cite{bao2026anchorreasoning}.
On NAVSIM, the three public planners are used with their released checkpoints, prompts, and image preprocessing.
Unless otherwise specified, we generate $K{=}16$ samples at temperature 1 without top-$k$/top-$p$ truncation.
The verifier uses the architecture described in \cref{sec:method} and is trained separately for each planner.

\begin{table*}[t]
\centering
\footnotesize
\setlength{\tabcolsep}{5pt}
\begin{tabular}{llccccccc}
\toprule
 & & \multicolumn{3}{c}{\valr} & \multicolumn{2}{c}{\valpool} & \multicolumn{2}{c}{\dev}\\
\cmidrule(lr){3-5}
\cmidrule(lr){6-7}
\cmidrule(lr){8-9}
Backbone & Selection
& \RFS$\uparrow$
& ADE$\downarrow$
& FDE$\downarrow$
& ADE$\downarrow$
& FDE$\downarrow$
& ADE$\downarrow$
& FDE$\downarrow$\\
\midrule

Cosmos-Reason2 (2B)
& Greedy
& \textbf{7.56}
& 3.06
& 7.58
& 2.06
& 5.58
& 2.26
& 6.24\\

& MBR sample
& 7.47
& 3.00
& 7.34
& 2.12
& 5.73
& 2.17
& 5.88\\

\oursrow
& \ours, $\lambda{=}0$
& 7.53
& \hl{2.74}
& 6.67
& 1.78
& 4.83
& 1.75
& 4.67\\

\oursrow
& \ours, $\lambda{=}1$
& 7.47
& 2.75
& \hl{6.59}
& \hl{1.74}
& \hl{4.68}
& \hl{1.71}
& \hl{4.57}\\

& \textcolor{gray}{Oracle (closest to GT)}
& \textcolor{gray}{7.79}
& \textcolor{gray}{1.43}
& \textcolor{gray}{3.31}
& \textcolor{gray}{0.90}
& \textcolor{gray}{2.38}
& \textcolor{gray}{0.90}
& \textcolor{gray}{2.37}\\

\midrule

Qwen3-VL (8B)
& Greedy
& 7.54
& 3.04
& 7.54
& 2.05
& 5.52
& 2.21
& 6.07\\

& MBR sample
& 7.39
& 3.01
& 7.39
& 2.09
& 5.68
& 2.17
& 5.87\\

\oursrow
& \ours, $\lambda{=}0$
& \hl{7.57}
& 2.69
& 6.44
& 1.84
& 4.92
& 1.81
& 4.72\\

\oursrow
& \ours, $\lambda{=}1$
& 7.49
& \hl{2.56}
& \hl{6.21}
& \hl{1.72}
& \hl{4.69}
& \hl{1.65}
& \hl{4.34}\\

& \textcolor{gray}{Oracle (closest to GT)}
& \textcolor{gray}{7.91}
& \textcolor{gray}{1.40}
& \textcolor{gray}{3.17}
& \textcolor{gray}{0.91}
& \textcolor{gray}{2.38}
& \textcolor{gray}{0.90}
& \textcolor{gray}{2.42}\\

\midrule

Alpamayo 1.5 (10B)
& Greedy
& 7.81
& 2.94
& 7.06
& 1.79
& 4.66
& 2.03
& 5.32\\

& MBR sample
& 7.73
& 2.76
& 6.45
& 1.85
& 4.82
& 1.79
& 4.43\\

\oursrow
& \ours, $\lambda{=}0$
& 7.86
& 2.59
& 6.11
& 1.61
& 4.18
& 1.52
& 3.82\\

\oursrow
& \ours, $\lambda{=}1$
& \hl{7.90}
& \hl{2.47}
& \hl{5.73}
& \hl{1.52}
& \hl{3.95}
& \hl{1.46}
& \hl{3.63}\\

& \textcolor{gray}{Oracle (closest to GT)}
& \textcolor{gray}{8.14}
& \textcolor{gray}{1.19}
& \textcolor{gray}{2.48}
& \textcolor{gray}{0.70}
& \textcolor{gray}{1.76}
& \textcolor{gray}{0.64}
& \textcolor{gray}{1.59}\\

\bottomrule
\end{tabular}

\caption{\textbf{Main results on WOD-E2E} ($K{=}16$).
ADE and FDE are measured in meters at 5\,s against the ground-truth future.
The oracle selects, among the greedy trajectory and 16 samples, the candidate closest to the ground-truth future.}
\label{tab:main}
\end{table*}

\subsection{Main results}
\label{sec:main_results}

\cref{tab:main} shows that \ours consistently improves trajectory accuracy across all three WOD-E2E planners.
With $\lambda{=}1$, \ours reduces both ADE and FDE on \valpool by about 15\% relative to greedy decoding across all three backbones, with similar improvements on \valr.
The consistent gains across model scales and pre-training indicate that the benefit is not specific to a particular VLM planner.
Increasing $\lambda$ places greater emphasis on agreement with the ground-truth future, improving ADE and FDE while slightly shifting \RFS.
In contrast, MBR selection lowers \RFS and increases ADE on \valpool for all three models, showing that sampling or consensus alone does not recover the same gain.

\begin{table*}[t]
\centering
\footnotesize
\setlength{\tabcolsep}{2.6pt}
\begin{tabular}{lccc|ccccccccccc}
\toprule
Test, 10B
& \RFS$\uparrow$
& ADE$_3\downarrow$
& ADE$_5\downarrow$
& Constr.
& Inters.
& Ped.
& Cycl.
& Multi-l.
& Single-l.
& Cut-in
& Debris
& Spec.\,veh.
& Spotl.
& Others\\
\midrule

Greedy
& 7.74 & 1.35 & 2.99
& 8.26 & 7.74 & 7.79 & 7.53
& \textbf{7.75} & 8.37 & 7.91
& 7.98 & \textbf{8.11} & 6.73 & 7.03\\

\oursrow \ours, $\lambda{=}1$
& \hl{7.81} & \hl{1.25} & \hl{2.68}
& \hl{8.38} & \hl{7.83} & \hl{7.84} & \hl{7.67}
& 7.72 & \hl{8.40} & \hl{7.96}
& \hl{8.02} & 8.06 & \hl{6.78} & \hl{7.19}\\

\bottomrule
\end{tabular}

\caption{\textbf{Official WOD-E2E test-server results}
(Alpamayo 1.5, 1,505 frames).
Both rows use the same checkpoint and differ only in trajectory selection.
All reported values are official WOD-E2E metrics.}
\label{tab:test}
\end{table*}

\begin{figure}[t]
\centering
\includegraphics[width=\columnwidth]{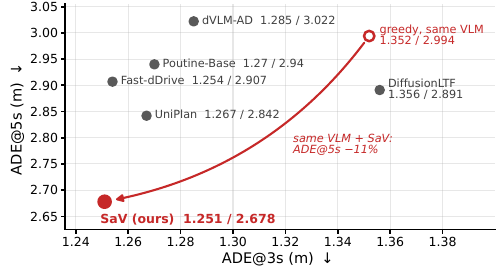}
\caption{\textbf{Official WOD-E2E test-set trajectory accuracy.}
ADE@3s and ADE@5s of published methods using the same three front cameras and without reinforcement-learning post-training.
Lower left is better; the arrow shows greedy decoding and \ours using the same Alpamayo-1.5 checkpoint.}
\label{fig:compare}
\end{figure}

\noindent\textbf{Official WOD-E2E test server.}
The validation gains persist on the official test set (\cref{tab:test}).
Using the same Alpamayo-1.5 checkpoint, \ours raises \RFS from 7.74 to 7.81 while reducing ADE@3s from 1.35 to 1.25\,m and ADE@5s from 2.99 to 2.68\,m.
The improvement spans 9 of the 11 scenario clusters, indicating that it is not driven by a single scenario type.
Among published methods using the same three front cameras and without reinforcement-learning post-training, \ours achieves the lowest ADE at both 3\,s and 5\,s (\cref{fig:compare}).
The shift from greedy decoding to \ours with the same checkpoint further isolates the gain to trajectory selection rather than additional planner training.

\begin{table}[t]
\centering
\footnotesize
\setlength{\tabcolsep}{2.6pt}
\renewcommand{\arraystretch}{1.05}

\begin{tabular}{@{}lccc@{}}
\toprule
Method
& \shortstack[c]{Curious-VLA\\SFT (3B)}
& \shortstack[c]{Curious-VLA\\RL (3B)}
& \shortstack[c]{ReCogDrive-VLM\\(8B)}\\
\midrule

& \multicolumn{3}{c}{\textit{PDMS} $\uparrow$}\\
\cmidrule(lr){2-4}

Greedy
& 87.00 & 89.90 & 85.84\\

Most likely sample
& 85.87 & 89.10 & 85.01\\

MBR sample
& 87.07 & 89.87 & 86.09\\

\oursrow \ours, ADE target
& 86.91 & 89.39 & 86.08\\

\oursrow \ours, PDMS target
& 87.01 & 89.63 & \hl{86.59}\\

\textcolor{gray}{Best candidate}
& \textcolor{gray}{95.49}
& \textcolor{gray}{96.06}
& \textcolor{gray}{94.78}\\

\midrule

& \multicolumn{3}{c}{\textit{ADE@4s (m)} $\downarrow$}\\
\cmidrule(lr){2-4}

Greedy
& 0.70 & 0.90 & 0.71\\

\oursrow \ours, ADE target
& \hl{0.63} & \hl{0.68} & \hl{0.64}\\

\bottomrule
\end{tabular}

\caption{\textbf{Cross-dataset results on NAVSIM}
(navtest, 12,146 scenes).
All planners use their released checkpoints.
\ours results are averaged over three verifier seeds.
}
\label{tab:navsim}
\end{table}

\noindent\textbf{Cross-dataset evaluation on NAVSIM.}
We further evaluate Curious-VLA-SFT, Curious-VLA-RL~\cite{chen2026curious}, and ReCogDrive-VLM-8B~\cite{li2026recogdrive} using their released checkpoints.
Across all three planners, the best of the greedy trajectory and 16 samples exceeds greedy decoding by 6.2--9.0 PDMS, while most-likely and MBR selection provide little or no benefit.
This confirms that the gap between trajectory generation and selection also exists beyond WOD-E2E.
With ground-truth supervision, \ours reduces ADE@4s by 9--25\% across all three planners.
PDMS gains are less consistent: using PDM score as the verifier target improves ReCogDrive-VLM-8B from 85.84 to 86.59, but brings little benefit to the two Curious-VLA variants.
Overall, these results show that the trajectory-selection gap generalizes across datasets and planners, while the effectiveness of verification depends on the supervision target.

\begin{figure*}[t]
\centering
\includegraphics[width=\textwidth]{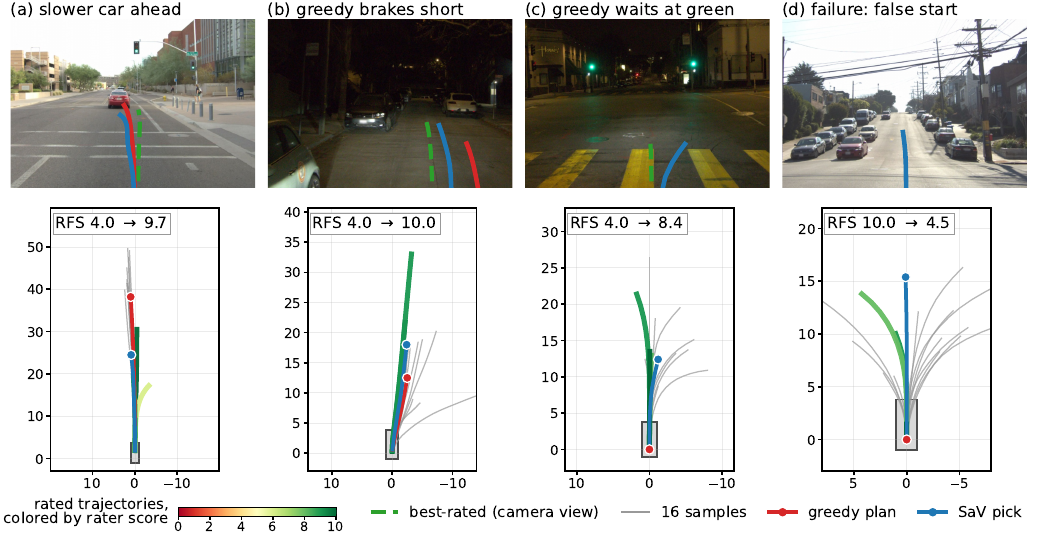}
\caption{\textbf{Qualitative results} (Alpamayo 1.5 on \valr).
(a) \ours maintains a larger gap to a slower vehicle ahead.
(b) \ours avoids the premature braking of the greedy plan.
(c) At a green light, greedy decoding remains stopped while \ours selects a moving trajectory.
(d) A failure case where \ours starts but the raters prefer to wait or turn left.}
\label{fig:qual}
\end{figure*}

\subsection{Qualitative results and failure cases}
\cref{fig:teaser,fig:qual} illustrate several types of errors corrected by \ours.
Beyond recovering missed starts at stop signs and green lights (\cref{fig:teaser}, left; \cref{fig:qual}c), \ours corrects both longitudinal and lateral planning errors: it maintains a safer gap to a slower vehicle (\cref{fig:qual}a), avoids unnecessary braking (\cref{fig:qual}b), and recovers the preferred direction when greedy decoding selects an incorrect turn (\cref{fig:teaser}, right).
In each case, the sampled set already contains a better trajectory, and the verifier selects it using trajectory and ego-state cues.

Failures can occur when the preferred maneuver depends on intent information unavailable to the verifier.
In \cref{fig:qual}d, for example, \ours selects a plausible straight-ahead start, while the raters prefer to wait or turn left.
This suggests that incorporating route or maneuver intent could help disambiguate such cases.

\subsection{Ablations}

\begin{table}[t]
\centering
\footnotesize
\setlength{\tabcolsep}{2.6pt}
\resizebox{\columnwidth}{!}{%
\begin{tabular}{lcccccc}
\toprule
& \multicolumn{2}{c}{Cosmos-Reason2}
& \multicolumn{2}{c}{Qwen3-VL}
& \multicolumn{2}{c}{Alpamayo 1.5}\\
\cmidrule(lr){2-3}
\cmidrule(lr){4-5}
\cmidrule(lr){6-7}
Verifier
& \RFS$\uparrow$ & ADE$\downarrow$
& \RFS$\uparrow$ & ADE$\downarrow$
& \RFS$\uparrow$ & ADE$\downarrow$\\
\midrule

\oursrow All features
& 7.47 & \hl{1.74}
& 7.49 & \hl{1.72}
& \hl{7.90} & \hl{1.52}\\

$-$ kinematics
& 7.39 & 1.90
& 7.38 & 1.95
& 7.75 & 1.75\\

$-$ ego state
& 7.56 & 2.00
& 7.47 & 2.00
& 7.73 & 1.74\\

$-$ candidate set
& 7.39 & 1.77
& 7.49 & 1.81
& 7.82 & 1.60\\

\quad $-$ likelihood only
& 7.46 & 1.74
& 7.52 & 1.77
& 7.90 & 1.58\\

\quad $-$ agreement only
& 7.46 & 1.74
& 7.54 & 1.81
& 7.85 & 1.57\\

\midrule
\multicolumn{7}{l}{\textit{Cross-backbone transfer}}\\

Cosmos-trained verifier
& -- & --
& 7.57 & 1.75
& 7.78 & 1.52\\

\bottomrule
\end{tabular}}
\caption{\textbf{Verifier ablations} ($\lambda{=}1$).
\RFS is evaluated on \valr and ADE@5s on \valpool.
Feature rows remove the indicated inputs and retrain the verifier.
The final row applies the Cosmos-2B verifier to the other backbones without retraining.}
\label{tab:features}
\end{table}

\noindent\textbf{What does the verifier rely on?}
Kinematic and ego-state features contribute most to trajectory selection (\cref{tab:features}).
Removing either increases ADE by 0.16--0.28\,m across the three backbones, reducing the gain over greedy decoding.
Candidate-set agreement and likelihood provide smaller complementary improvements.
These results indicate that the verifier primarily evaluates whether a candidate is dynamically consistent with the current ego motion, while information from the candidate distribution further refines the selection.

\noindent\textbf{Does the verifier transfer across backbones?}
A verifier trained only on Cosmos-Reason2 candidates transfers directly to Qwen3-VL and Alpamayo-1.5 without retraining, reaching 1.75 and 1.52\,m ADE on \valpool, compared with 1.72 and 1.52\,m for their model-specific verifiers (\cref{tab:features}).
The small gap suggests that much of the learned trajectory-selection signal is shared across VLM backbones rather than being specific to one planner.

\begin{table}[t]
\centering
\footnotesize
\setlength{\tabcolsep}{2.7pt}
\begin{tabular}{lrcccccc}
\toprule
& & \multicolumn{2}{c}{Cosmos-Reason2}
& \multicolumn{2}{c}{Qwen3-VL}
& \multicolumn{2}{c}{Alpamayo 1.5}\\
\cmidrule(lr){3-4}
\cmidrule(lr){5-6}
\cmidrule(lr){7-8}
GT behavior & $n$
& Greedy & \ours
& Greedy & \ours
& Greedy & \ours\\
\midrule

waits
& 238
& \textbf{0.21} & 0.50
& \textbf{0.18} & 0.42
& \textbf{0.23} & 0.51\\

starts
& 119
& 3.12 & \hl{2.49}
& 3.14 & \hl{2.40}
& 3.05 & \hl{2.22}\\

stops
& 163
& \textbf{1.45} & 1.53
& 1.55 & \hl{1.47}
& 1.38 & \hl{1.34}\\

decelerates
& 348
& 1.93 & \hl{1.85}
& 2.03 & \hl{1.83}
& 1.63 & \hl{1.53}\\

keeps speed
& 587
& 1.62 & \hl{1.37}
& 1.56 & \hl{1.40}
& 1.56 & \hl{1.30}\\

accelerates
& 461
& 3.61 & \hl{2.64}
& 3.55 & \hl{2.65}
& 2.82 & \hl{2.19}\\

\midrule

all
& 1,916
& 2.06 & \hl{1.74}
& 2.05 & \hl{1.72}
& 1.79 & \hl{1.52}\\

\bottomrule
\end{tabular}
\caption{\textbf{ADE@5s by ground-truth longitudinal behavior} on \valpool ($\lambda{=}1$).
Wait/start denote stationary initial states; the remaining groups are defined by the change in ego speed over the prediction horizon.}
\label{tab:behavior}
\end{table}

\noindent\textbf{Where does the gain come from?}
The largest improvements occur when the ground-truth future requires continued progress (\cref{tab:behavior}).
ADE decreases by 20--27\% when the ego starts from rest and by 22--27\% during acceleration, with consistent gains when maintaining speed.
In contrast, ADE increases substantially on waiting frames because the verifier sometimes selects a creeping or starting candidate. This is reflected in the false-start rate (\cref{tab:fts}), showing that SaV partly trades failed starts for false starts. Overall, the largest gains come from correcting the inertia of greedy decoding, particularly missed starts and insufficient progress during acceleration, rather than uniformly improving all driving decisions.

\begin{table}[t]
\centering
\footnotesize
\setlength{\tabcolsep}{4pt}
\begin{tabular}{lccc}
\toprule

\multicolumn{4}{l}{\textit{Target weight} $\lambda$
(Cosmos-Reason2, $K{=}16$)}\\
Setting
& \valr \RFS$\uparrow$
& \valr ADE$\downarrow$
& \valpool ADE$\downarrow$\\
\midrule

Greedy
& \textbf{7.56} & 3.06 & 2.06\\

$\lambda{=}0$
& 7.53 & 2.74 & 1.78\\

$\lambda{=}0.25$
& 7.51 & 2.78 & 1.78\\

$\lambda{=}0.5$
& 7.46 & 2.77 & 1.75\\

$\lambda{=}1$
& 7.47 & 2.75 & \hl{1.74}\\

$\lambda{=}2$
& 7.41 & \hl{2.67} & \hl{1.74}\\

\midrule

\multicolumn{4}{l}{\textit{Number of samples} $K$
(\valpool ADE$\downarrow$, $\lambda{=}1$)}\\
Setting
& Cosmos-Reason2
& Qwen3-VL
& Alpamayo 1.5\\
\midrule

Greedy
& 2.06 & 2.05 & 1.79\\

$K{=}4$
& 1.86\,{\scriptsize$\pm$.01}
& 1.87\,{\scriptsize$\pm$.01}
& 1.68\,{\scriptsize$\pm$.04}\\

$K{=}8$
& 1.78\,{\scriptsize$\pm$.01}
& 1.87\,{\scriptsize$\pm$.07}
& 1.55\,{\scriptsize$\pm$.01}\\

$K{=}16$
& \hl{1.73}\,{\scriptsize$\pm$.00}
& \hl{1.73}\,{\scriptsize$\pm$.01}
& \hl{1.52}\,{\scriptsize$\pm$.01}\\

\bottomrule
\end{tabular}
\caption{\textbf{Target and sampling ablations.}
The lower block reports mean $\pm$ std over four verifier seeds.}
\label{tab:lambda}
\end{table}

\noindent\textbf{Training target.}
The target weight $\lambda$ controls the balance between the trust-region score and geometric agreement with the ground-truth trajectory (\cref{tab:lambda}).
Increasing $\lambda$ improves ADE but gradually lowers \RFS.
On \valpool, $\lambda{=}1$ reaches the same 1.74\,m ADE as $\lambda{=}2$ while retaining a higher \RFS on \valr, motivating $\lambda{=}1$ as the default setting.

\noindent\textbf{Number of samples.}
Larger candidate sets generally improve trajectory selection (\cref{tab:lambda}), with $K{=}16$ giving the lowest mean ADE for all three backbones.
Most of the gain is already recovered with fewer samples for Cosmos-Reason2 and Alpamayo 1.5, whereas Qwen3-VL shows greater variation across verifier seeds.

\begin{figure}[t]
\centering
\includegraphics[width=\columnwidth]{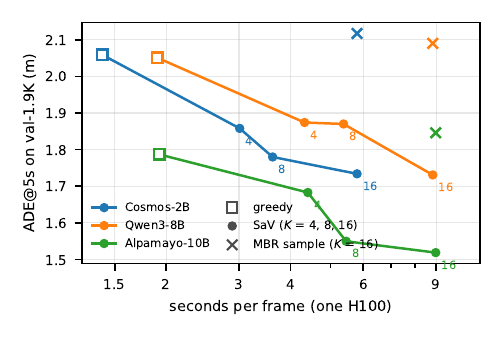}
\caption{\textbf{Accuracy--cost trade-off.}
ADE@5s on \valpool versus measured inference time on one H100 GPU for greedy decoding, \ours with $K{=}4,8,16$, and MBR with $K{=}16$.}
\label{fig:cost}
\end{figure}

\noindent\textbf{Inference cost.}
The accuracy gain comes primarily from generating additional candidates. Relative to greedy decoding, SaV requires approximately 2.2--2.3$\times$ the inference time with $K{=}4$, 2.6--2.8$\times$ with $K{=}8$, and 4.1--4.7$\times$ with $K{=}16$ on one H100 GPU. Feature extraction and verifier scoring add only about 15\,ms, so candidate generation dominates the overhead. The current $K{=}16$ configuration is therefore not intended for real-time deployment. Smaller candidate sets recover a substantial fraction of the accuracy gain, suggesting that adaptive or selective candidate generation could provide a better accuracy--latency trade-off.

\begin{figure}[t]
\centering
\includegraphics[width=\columnwidth]{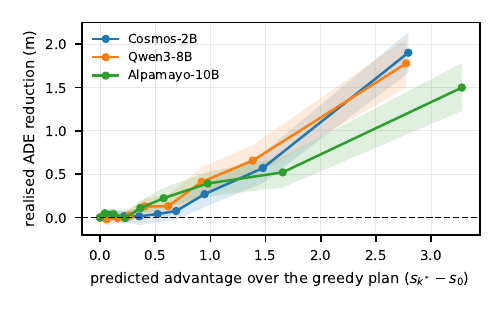}
\caption{\textbf{Predicted advantage tracks realized improvement.}
\valpool frames are binned by the predicted score advantage of the selected candidate over greedy decoding and plotted against the realized ADE@5s reduction (mean with 95\% bootstrap interval; $\lambda{=}1$).}
\label{fig:calib}
\end{figure}

\noindent\textbf{Can the verifier identify when to replace greedy decoding?}
The score advantage of the selected candidate over the greedy trajectory is positively associated with the realized ADE improvement (\cref{fig:calib}).
Across all three backbones, larger predicted advantages generally correspond to larger realized gains, supporting the use of a margin to selectively retain the greedy trajectory when the verifier has low confidence.
\section{Conclusion}

We identify trajectory selection as a source of error separate from trajectory generation in autoregressive VLM driving planners. Across the WOD-E2E and NAVSIM, greedy decoding can miss better trajectories already present in the model distribution. We therefore introduce SaV, which keeps the planner frozen and selects among sampled trajectories with a lightweight learned verifier. The verifier ranks candidates using trajectory kinematics, candidate-set agreement and likelihood, and ego-motion cues, with supervision from ground-truth. On WOD-E2E, SaV reduces ADE and FDE by about 15\% across three VLM planners on the validation set. RFS changes are comparatively small, with slight decreases for two planners on val-456 and an improvement from 7.74 to 7.81 for Alpamayo-1.5 on the official test set, where ADE@5s decreases by 11\%. On NAVSIM, SaV reduces ADE@4s by 10--24\%, while PDMS changes are less consistent across planners. Overall, the results show that trajectory selection can improve displacement accuracy without retraining the underlying planner.

\noindent\textbf{Limitations.}
SaV introduces substantial inference overhead because multiple VLM trajectories must be generated, and the current $K{=}16$ configuration is not real-time. Reducing the sampling budget or generating additional candidates only when needed may improve practical efficiency. Our evaluation is also limited to open-loop trajectory selection; improvements in displacement error or rated trajectory quality do not necessarily translate to better interactive closed-loop behavior. Evaluating SaV under reactive closed-loop settings is an important direction for future work.

{
    \small
    \bibliographystyle{ieeenat_fullname}
    \bibliography{main}
}
\end{document}